\pdfoutput=1
\documentclass[11pt]{article}

\usepackage[final]{acl}
\usepackage{hyperref}
\usepackage{tabularx}
\usepackage{booktabs}

\usepackage{siunitx}

\usepackage{multirow}

\usepackage{times}
\usepackage{latexsym}

\usepackage[T1]{fontenc}
\usepackage[utf8]{inputenc}

\usepackage{microtype}

\usepackage{inconsolata}

\usepackage{graphicx}
\usepackage{makecell}  % For \makecell

\usepackage{xcolor} 
\usepackage{amssymb}
\usepackage{xspace}

\usepackage{soul}
\usepackage{fancyvrb}
\usepackage{tcolorbox}

\definecolor{titlebar}{RGB}{50,50,50}
\definecolor{promptbox}{RGB}{230,240,255}
\definecolor{outputbox}{RGB}{255,230,240}
\definecolor{docbox}{RGB}{230,255,240}

\definecolor{repetition}{RGB}{255,0,0}

\title{Low-Perplexity LLM-Generated Sequences and Where To Find Them}

\author{
  \textbf{Arthur Wuhrmann\textsuperscript{1}},
  \textbf{Anastasiia Kucherenko\textsuperscript{2}}, 
  \textbf{Andrei Kucharavy\textsuperscript{3}}
\\
  \textsuperscript{1}École Polytechnique Fédérale de Lausanne, Switzerland\\
  \textsuperscript{2}Institute of Entrepreneurship and Management, HES-SO Valais-Wallis, Switzerland \\
  \textsuperscript{3}Institute of Informatics, HES-SO Valais-Wallis, Switzerland
\\
  \small{
    \textbf{Correspondence:} \texttt{\href{mailto:arthur.wuhrmann@epfl.ch}{arthur.wuhrmann@epfl.ch}}
    }
}

\begin{document}
\maketitle
\begin{abstract}
As Large Language Models (LLMs) become increasingly widespread, understanding how specific training data shapes their outputs is crucial for transparency, accountability, privacy, and fairness. To explore how LLMs leverage and replicate their training data, we introduce a systematic approach centered on analyzing low-perplexity sequences—high-probability text spans generated by the model. Our pipeline reliably extracts such long sequences across diverse topics while avoiding degeneration, then traces them back to their sources in the training data. Surprisingly, we find that a substantial portion of these low-perplexity spans cannot be mapped to the corpus. For those that do match, we quantify the distribution of occurrences across source documents, highlighting the scope and nature of verbatim recall and paving a way toward better understanding of how LLMs training data impacts their behavior.
\end{abstract}
\section{Introduction}

While Large Language Models (LLMs) are increasingly applied across various domains, the ways in which they leverage their training data during inference remains only partially understood~\cite{MIT2024LLMs,10.1145/3442188.3445922, liang2024mappingincreasingusellms}. Research on training data attribution (TDA) in LLMs~\cite{carlini2021extractingtrainingdatalarge, cheng2025trainingdataattributiontda} aims to answer this question, but identifying which specific parts of the data contribute to a model's output. TDA is considered essential for enhancing transparency, effective debugging, accountability, and addressing concerns related to privacy and fairness in LLMs~\cite{cheng2025trainingdataattributiontda, akyurek-etal-2022-towards, olmotrace}. %\ana{add conclusive phrase for explainable AI}

Currently, there are two principal approaches for TDA - causal and similarity-based. Causal TDA uses direct experimental methods such retraining and gradient-based techniques that quantify the precise causal contribution of individual training samples to model outputs~\cite{guu2023simfluencemodelinginfluenceindividual, DBLP:journals/corr/abs-2310-00902, detecting_Yijun, akyurek-etal-2022-towards, chang2024scalableinfluencefacttracing, wu-etal-2024-enhancing-training}. While offering theoretical guarantees about causality, their computational cost increases dramatically with model size, making them infeasible in practice.

Similarity-based TDA~\cite{olmotrace, carlini2021extractingtrainingdatalarge, khandelwal2020generalizationmemorizationnearestneighbor, deguchi2025softmatchasoftfastpattern} identifies training samples that resemble model outputs, assuming similar content likely influenced generation. While similarity does not guarantee causal influence and this attribution is approximate, this approach is computationally efficient and scales well to large models, making it feasible in practice. Similarity-based TDA includes approaches such as nearest-neighbor searches in embedding spaces and exact string matching for verbatim recall. In this paper, we focus on the latter, which connects to the established field of novelty~\cite{mccoy-etal-2023-much, merrill2024evaluatingngramnoveltylanguage} and memorization in LLMs~\cite{carlini2023quantifyingmemorizationneurallanguage, Al_Kaswan_2024, carlini2023extractingtrainingdatadiffusion, 10.5555/3495724.3495966, prashanth2025recitereconstructrecollectmemorization}, studying instances where models produce verbatim recall of training data. Recently, the first tool for efficient TDA based on exact memorization was introduced~\cite{olmotrace}, underscoring the practical importance of such approaches.

In this paper, we study how low-perplexity sequences in LLM-generated output are connected to its verbatim recall. Perplexity is a standard metric used to evaluate a model's ability to predict tokens, with lower perplexity indicating higher confidence in its predictions. It is widely employed for model evaluation, fine-tuning, comparison and assessing text generation quality. In the context of training data attribution (TDA), there is a hypothesis that long low-perplexity sequences suggest either degeneration or verbatim copying from the training data~\cite{gao2019representationdegenerationproblemtraining, prashanth2025recitereconstructrecollectmemorization}. We aim to empirically test this statement, while proposing a method to better understand LLMs' verbatim recall through low-perplexity analysis. 

We present an open-source pipeline\footnote{The code is available at \url{https://github.com/Reliable-Information-Lab-HEVS/HAIDI-Graphs}} designed to identify and trace low-perplexity spans in LLM outputs. By targeting specialized domains with rich, distinctive terminology, our approach efficiently extracts long, low-perplexity segments suitable for in-depth analysis. These segments are then mapped back to their origins using indexing and search tools. Although we experimented with both the well-established Elasticsearch~\cite{gormley2015elasticsearch} and the recently emerged state-of-the-art Infinigram~\cite{infinigram}, we report only Infinigram results due to its superior scalability and efficiency for large-scale mapping.

%Our analysis reveals that many of these low-perplexity spans cannot be matched to the training data. For those that do, we further categorize the types of memorization behaviors involved. This classification enables us to accurately quantify their distribution across various specialized topics, offering deeper insights into how LLMs recall and replicate information.

Our analysis provides deeper insights into how LLMs recall and replicate information. First, we observe that results vary depending on the topic of LLM input, its representation in the training data, and its degree of specialization. Second, we find that a significant portion of low-perplexity spans, ranging from $30\%$ to $60\%$, cannot be matched to the training data. For those that can be matched, we further categorize different types of memorization behaviors, noting that verbatim recall can arise for various reasons. Finally, this classification allows us to quantify that approximately $20\%$ of low-perplexity spans correspond to a number of documents small enough for manual review.

%\subsection*{Training Data Indexing} A lot of indexing technique emerged for the training data of LLMs, trying to improve data transparency, copyright checking, and privacy. These techniques include notably Data portraits~\cite{marone2023dataportraits}, Infinigram~\cite{liu2025infinigramscalingunboundedngram} or WIMBD~\cite{elazar2024whatsbigdata}. 
%In our investigation, since we need to count the number of occurences in the training data, BF cannot be used. While WIMBD could have been used, the indexing of the Pile is not accessible, and Infinigram proposes an efficient API, which is the reason we chose it. Infinigram works by constructing suffixes and indexes them. The tool can count the number of occurences of any text in a pre-indexed training data, and recover the documents that contain this text. \art{should we cite GSM-symbolic ? should we explain in more details how their techniques work.} \andrei{We should, and should be a bit more specific about what it does. We can say we chose Infinigram as well due to its efficiency}.

\begin{figure}[h]  % 't' places it at the top of the page
    \centering
    \includegraphics[width=\columnwidth]{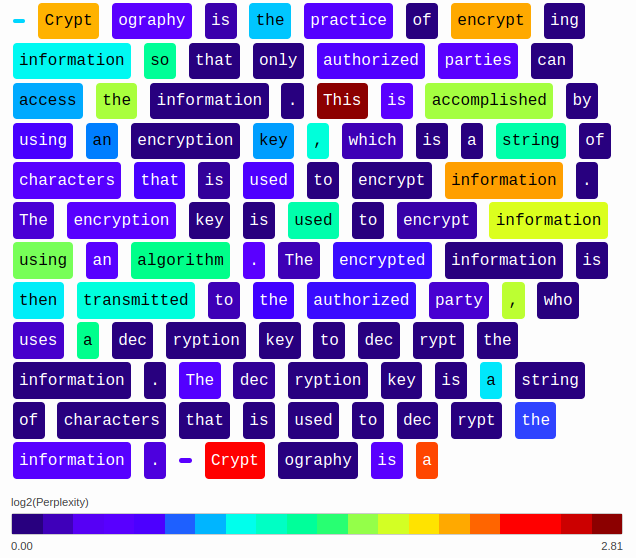}
    \caption{Visualization of a generated subsequence that contains two different low-perplexity sequences longer than 5 tokens. We have \texttt{decryption key to decrypt the information} and \texttt{string of characters that is used to decrypt}. Both having 9 tokens, they will be split in $9+1-6=4$ windows of 6-contiguous tokens each.}
    \label{fig:sample-viz}
\end{figure}

\section{Experimental setup} 
\label{sec:setup}
\subsection*{LLM model and training data}
To study low-perplexity sequences we use 
the Pythia model~\cite{biderman2023pythiasuiteanalyzinglarge} with size of 6.9 billion parameters trained on \emph{The Pile}~\cite{gao2020pile800gbdatasetdiverse}, which transforms into 300 billion tokens using Pythia tokenizer~\cite{biderman2023pythiasuiteanalyzinglarge}, with a vocabulary size $|V| = 50,254$.%\art{manually verified}.

\subsection*{Choosing topics and prompts}
To follow our goal of finding low-perplexity sequences, we focus on keyword-specific topics for this study. Therefore, we choose \textbf{genetics, nuclear physics, drugs, and cryptography}, specialized domains in which the team has experience to verify the validity of LLM outputs. Since we work with The Pile dataset, those topics are represented at least as part of its Wikipedia subset. %\art{Therefore, we use the Wikipedia API to obtain quotes from its 2020 version, when the Pile was created. We moreover verify that these quotes belonged to the Pile using Infinigram~\cite{infinigram}.}

In total, for each topic, we select 40 articles from the Wikipedia version included in the Pile and extract a random quote consisting of 20 to 40 tokens. This quote serves as a prompt for the Pythia model to complete and extend. For each prompt we run $5$ generations to average the results. This approach provides $200$ prompts per topic and $800$ prompts in total. 

\subsection*{LLM output generation and perplexities}
LLMs generate output sequentially---token by token---by sampling the next token based on its logits values and key parameters: $\textrm{top}_k$, which restricts choices to the top $k$ most probable words; $\textrm{top}_p$, which selects the smallest set of words with a cumulative probability of $p$; and temperature $T$, which controls randomness. We set $\textrm{top}_k = 20$, $\textrm{top}_p = 0.8$, and $T = 0.7$, with alternative configurations discussed in Sec.~\ref{sec:temperature}.

The exact definition of the generation probability of each token ($x_i$) based on the previous tokens ($x_{<i}$) is   
\[
p(x_i|x_{<i}) = \frac{\exp(z_i/T)}{\sum_{j=1}^{|V|} \exp(z_j/T)},
\]
where $z_i$ are the raw logits and $|V|$ is the vocabulary size of the model.  
Then, the \emph{token perplexity} is: 
\begin{equation}\label{eq:perplexity}
    P(x_i) = \frac{1}{p(x_i|x_{<i})}.
\end{equation}
We define a \textbf{low-perplexity sequence} as a contiguous part of the LLM output where \textit{each token has a perplexity threshold $\log_2(P) \leq 0.152$ in base 2}, corresponding to a \textit{probability threshold of $0.9$ or higher}. These sequences have different lengths, so to compare the matches in the training data, we focus on their fixed-size subsequences. We call those \textbf{low-perplexity windows} and focus our choice on size of $6$ tokens. The choice of a 6-token window is justified as it is short enough to capture meaningful low-perplexity spans while being long enough to avoid random matches. Fig.~\ref{fig:sample-viz} shows a visualization of the generated tokens and perplexities values.

\subsection*{Matching to the training data and its quality}
Finally, we map low-perplexity windows to the training data. To achieve this, we use Infinigram~\cite{infinigram}. Once a low-perplexity window is matched to the training data, we estimate the significance of its text. We do this using perplexity values (as defined in Equation~\ref{eq:perplexity}), this time without additional context (i.e., tokens preceding the window), which is also known as \textit{standalone perplexity}. We denote it as \[
\hat{P}(x_k, \dots, x_{k+n}) = \textstyle 2^{-\frac{1}{n} \sum_{i=k}^{k+n} \log_2 p(x_i \mid [x_k, \dots, x_{i-1}])}
\] 
Low standalone perplexity indicates that the generated text is fluent, coherent, and resembles human-written language~\cite{gonen2024demystifyingpromptslanguagemodels}.%\art{is it okay to cite this? cant find much better}\andrei{good enough here}
\section{Results}
\subsection{Descriptive analysis of low-perplexity windows} 
We begin by identifying all low-perplexity sequences across the four chosen topics. The warm-up statistics in Table~\ref{tab:spans-stats} show that the average lengths of these sequences do not vary significantly between topics, and our choice of a fixed window size of $6$ is sufficiently modest.

\begin{table}[h]
\centering
\resizebox{0.5\columnwidth}{!}{%
\begin{tabular}{l|rr}
\toprule
Topic &  $\bar{L}$ & $\sigma_{L}$ \\
\midrule
Crypt2ography &  12 & 11 \\
Drugs &  14 & 15 \\
Genetics &  14 & 14 \\
Nuclear physics &  13 & 12 \\
\bottomrule
\end{tabular}%
}
\caption{ $\bar{L}$ (resp. $\sigma_{L}$) represents the average (resp. standard deviation) of the token lengths for low-perplexity sequences with at least 6 tokens.}
\label{tab:spans-stats}
\end{table}

\begin{figure}[h]  % 't' places it at the top of the page
    \centering
    \includegraphics[width=\columnwidth]{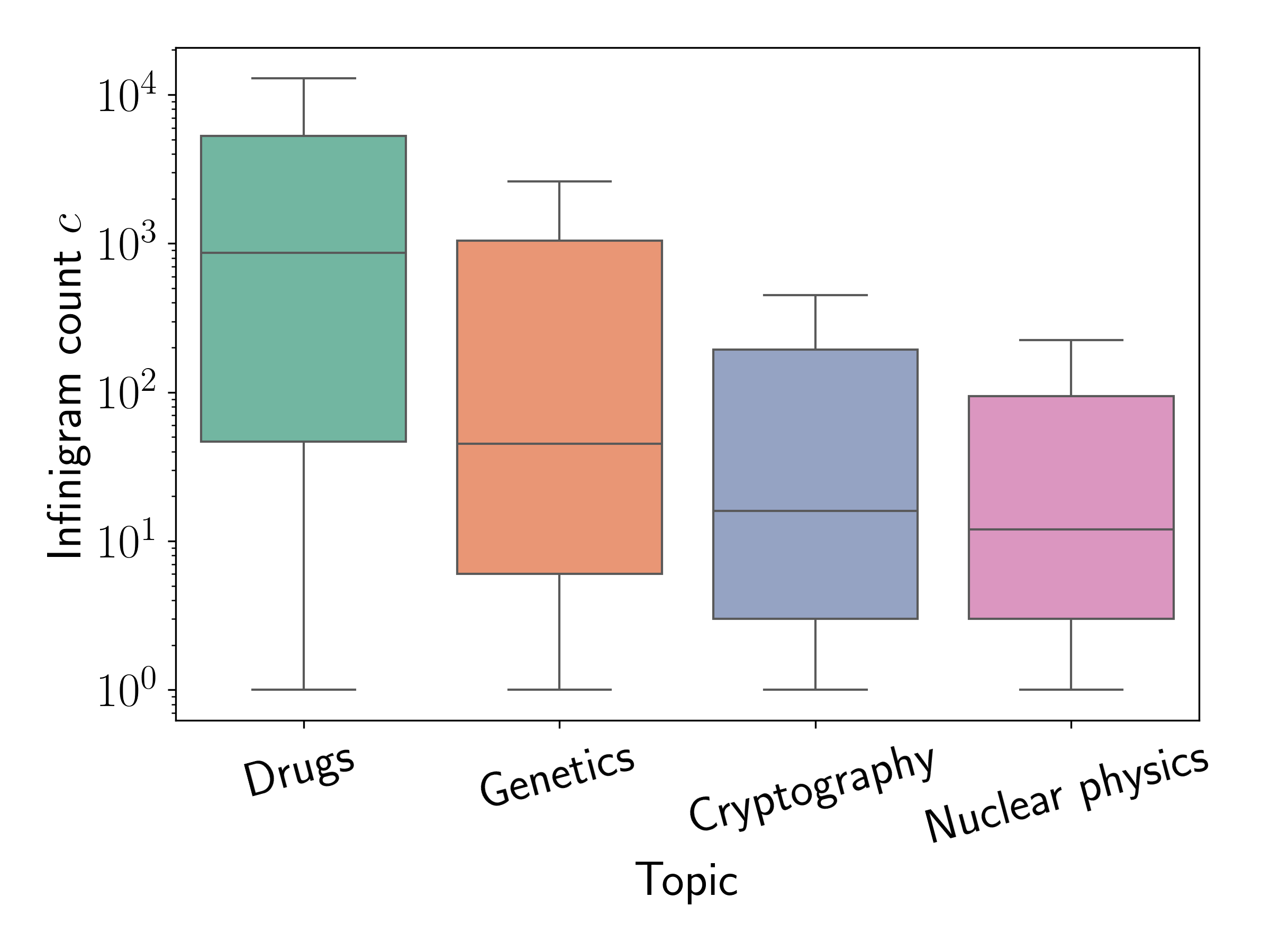}
    \caption{Boxplots comparing the number of matches of low-perplexity windows that occur in the training data, across different topics. %Given the log-scale, sequences with zero matches are excluded, but can be seen in Tab~\ref{tab:dataset-stats}.
    }
    \label{fig:box-plots}
\end{figure}

From selected low-perplexity sequences, we pass a sliding window of 6 tokens and stride 1 and proceed to our main interest -- low-perplexity windows matched to the training data. We denote the number of occurances by $c$. Figure~\ref{fig:box-plots} presents the comparison of windows at least with one match across different topics.  We observe having significantly more of long low-perplexity sequences on drugs. We believe this is due to the presence of repetitive long drug names and their strong connection to biomedical literature, which is widely represented in the Pile dataset through the inclusion of PubMed. On the other side, it is likely that nuclear physics is less present in the Pile, which explains the lower number of counts.

Above, only windows with at least one exact match in the training data are considered. While one might expect low-perplexity windows to almost always have matches, we verify this experimentally (Table~\ref{tab:sequence-stats}). Surprisingly, \emph{only 40\% of low-perplexity windows have at least one exact match ($N_{c>0}$)}. We also observe varying match counts across topics, likely due to differences in their specialization and corpus representation.

\begin{table}[h]
\centering
\resizebox{\columnwidth}{!}{%
\begin{tabular}{l|rrrrrr}
\toprule
Topic & {$N$} & {$N_{c>0}$} & {$N_{c>0}/N$} & {$N_\textrm{rep}/N$}  \\
\midrule
Cryptography & 1336 & 505 & 38\% & 32\% \\
Drugs & 988 & 659 & 67\% & 7.9\% \\
Genetics & 1337 & 481 & 36\% & 29\% \\
Nuclear physics & 1040 & 264 & 25\% & 15\% \\
Total & 4701 & 1909 & 41\% & 21\% \\

\bottomrule
\end{tabular}%
}
\caption{The total number of low-perplexity windows $N$ for each topic, number and percentage of those windows that have exact matching the training data $N_{c>0}$. $N_\textrm{rep}/N$ is the percentage of low-perplexity sequences repeating the prompt (see Appendix \ref{appendix:repetition}). }
\label{tab:sequence-stats}
\end{table}

Finally, examining the matched windows, we find that a significant fraction partially repeats the prompt ($N_\textrm{rep}$). We suspect this is due to the specialized keywords in the prompt and therefore we retain these repetitions for further analysis. Appendix \ref{appendix:repetition} presents an example of such repetition.
%and their properties do not significantly vary depending on the temperature of the generation.% For this, we ran the same experiments on 20 prompts, 5 generations, of the Genetics sequence for temperatures values ranging from 0.2 to 0.7.

\subsection{The nature of low-perplexity sequences }
\label{sec:categories}
\begin{figure}[h]  
    \centering
    \includegraphics[width=\columnwidth]{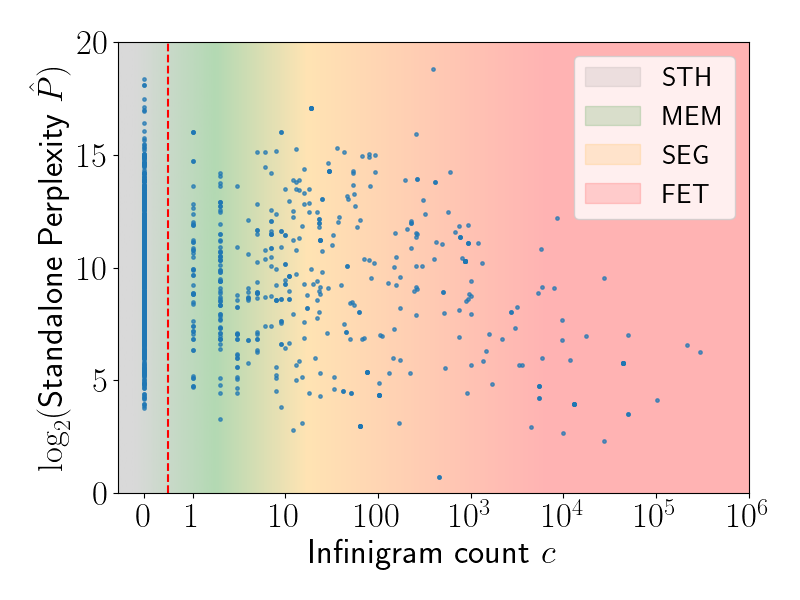}
    \caption{Illustration of the low-perplexity sequences, for the Cryptography topic.}
    \label{fig:cyber-scatter}
\end{figure}

Using two additional measures, we explore the behaviors exhibited by the model when generating low-perplexity sequences (Figure~\ref{fig:cyber-scatter}). First, we revisit the concept of stand-alone perplexity to assess how human-like the generated text appears.
Second, we categorize the low-perplexity windows into four groups based on their number of matches in the training data $(c)$, reflecting different recall and generalization behaviors. Since these behaviors can overlap, the group boundaries are not sharply defined. Therefore, in Figure~\ref{fig:cyber-scatter}, we intentionally use a color gradient to illustrate the smooth transition between categories. While we indicate specific thresholds for the match count $c$ below, these values are adjustable and intended to aid interpretation rather than impose strict divisions. Particular examples of each behavior can be found in Appendix~\ref{appendix:categories}.

\begin{itemize}
    \item \textbf{Synthetic coherence ($c=0$):} These windows are synthetically generated by the model without any exact matches in the training data. Interestingly, the stand-alone perplexities vary widely, including high values. However, as shown in Appendix~\ref{appendix:categories}, even the generations with the highest perplexity scores remain coherent and are not non-sensical.

    \item \textbf{Memorization $(0<c<5)$} The model has generated text containing highly specific knowledge, which can be traced back with high precision to its origins in the training data. Such traceability is particularly valuable for identifying instances of private and sensitive data leakage, memorized and reproduced by the model. An example is given in Appendix~\ref{appendix:match}.

    \item \textbf{Segmental replication $(5 \leq c < 50)$} These windows contain relatively niche information that appears across multiple sources, often reflecting standardized phrases or terminology within specific domains. Alongside memorization, segmental replication helps efficiently trace LLM outputs to their origins, revealing how specialized knowledge is represented. 
    
    \item \textbf{Frequently encountered text $ ( 50<c )$} These windows correspond to common phrases or widely used expressions that appear frequently across many documents in the training data. When $c$ becomes very large, it typically reflects standardized text such as legal disclaimers, licensing terms or HTML tags (i.e., \texttt{<div><\textbackslash div>)}, indicating heavy repetition across the corpus.
    %"Source Code Form is subject to the terms of the Mozilla Public License, v. 2.0. If a copy of the MPL was not distributed with this file, You can obtain one at http://mozilla.org/MPL/2.0/."
\end{itemize}

While the thresholds of $5$ and $50$ were chosen arbitrarily, fixing them enables consistent counting and comparison across topics, as shown in Table~\ref{tab:category-distribution}. Notably, around $20\%$ of low-perplexity windows fall into the memorization and segmental replication categories, matching to a number of documents small enough to be manually reviewed.

\begin{table}[h]
\setlength{\tabcolsep}{3pt}
\centering
\begin{tabular}{l|rrrrr}
\toprule {Topic} & {STH} & {MEM} & {SEG} & {FET} \\
\midrule
Cryptography & 62\% & 11\% & 13\% & 14\% \\
Drugs & 33\% & 7.5\% & 9.3\% & 50\% \\
Genetics & 64\% & 7.7\% & 11\% & 17\% \\
Nuclear physics & 75\% & 8.1\% & 9.3\% & 8\% \\
\bottomrule
\end{tabular}
\caption{Distribution of categories across topics. Categories: Synthetic coherence (STH), Memorization (MEM), Segmental replication (SEG), and  Frequently encountered text (FET).}\label{tab:category-distribution}
\end{table}

\subsection{LLM size and its generation parameters}
\label{sec:temperature}

In the previous experiments, we used the Pythia-$6.9B$ model with fixed generation parameters, as described in Section~\ref{sec:setup}. In this section, we repeat the experiments with alternative model settings and justify our initial choice.

First, we replicate the experiments across the Pythia model scaling suite (Table~\ref{tab:size-stats}). As model size increases, we observe a clear drop in both the number of low-perplexity windows and their matches to the training data. This supports our choice of the 6.9B model, which offers more meaningful responses, while any matching results would only improve in smaller models.

\begin{table}[h]
\centering
%\resizebox{\columnwidth}{!}{%
\begin{tabular}{l|rrrrr}
\toprule
Size & {$N$} & {$N_{c>0}$} & {$N_{>0}/N$} & $N_\textrm{rep}$ & $\hat{P}$ \\
\midrule
70M & 8528 & 2874 & 34\% & 118 & 9.2 \\
160M & 3676 & 1306 & 36\% & 428 & 8.4 \\
410M & 2274 & 716 & 31\% & 470 & 8.4 \\
1B & 2766 & 878 & 32\% & 752 & 8.6 \\
1.4B & 2123 & 673 & 32\% & 334 & 8.2 \\
2.8B & 1714 & 488 & 28\% & 402 & 8.6 \\
6.8B & 1337 & 481 & 36\% & 386 & 8.5 \\
\bottomrule
\end{tabular}%
\caption{Number of low-perplexity sequences and matches when varying the model sizes. Done on the Genetics topic.}
\label{tab:size-stats}
\end{table}

Further, we study the impact of varying the temperature parameter, which controls the LLM generation randomness (Table~\ref{tab:dataset-stats}).

\begin{table}[h]
\centering
%\resizebox{\columnwidth}{!}{%
\begin{tabular}{l|rrrrr}
\toprule
$T$ & {$N$} & {$N_{c>0}$} & {$N_{>0}/N$} & $N_\textrm{rep}$ & $\hat{P}$ \\
\midrule
0.2 & 8787 & 2908 & 33\% & 743 & 8.7 \\
0.3 & 6127 & 1918 & 31\% & 589 & 8.5 \\
0.4 & 4523 & 1461 & 32\% & 598 & 8.9 \\
0.5 & 3297 & 1091 & 33\% & 560 & 8.8 \\
0.6 & 1913 & 659 & 34\% & 310 & 8.6 \\
0.7 & 1337 & 481 & 36\% & 386 & 8.5 \\
\bottomrule
\end{tabular}%
\caption{Number of low-perplexity sequences and matches when varying the temperature. Done on the Genetics topic.}
\label{tab:dataset-stats}
\end{table}
 Lower temperature makes the model more deterministic, favoring high-probability tokens. We observe that it leads to a greater number of low-perplexity windows, however increases degeneration and more repetitive patterns in the LLM outputs. Also, interestingly, the overall percentage of non-zero matches, as well as the stand-alone perplexity, remains largely unchanged. These results explain our preference for a temperature value of $0.7$ — it provides a meaningful number of low-perplexity windows for analysis while reducing the extent of repetition.

\section{Conclusion}
We proposed a pipeline to identify and analyze low-perplexity sequences in LLM outputs. We categorized sequences by their match frequency in the training data and identified four distinct behaviors. We also conducted a statistical analysis of these categories, notably finding that many low-perplexity sequences do not match the corpus at all. This approach improves understanding of how models recall learned information and, in some cases, enables more efficient training data attribution.

% Long conclusion:
%To better understand how these sequences relate to the training data, we introduce a categorization scheme based on their frequency of matches in the corpus, resulting in four distinct behaviors. We further provide a statistical analysis of how often each behavior occurs across different specialized topics. This categorization not only sheds light on how models recall and replicate learned information—ranging from specialized domain knowledge to potentially private content—but also enables more efficient training data attribution in certain cases. By focusing on low-perplexity windows instead of full-text matches,

\clearpage

\section{Limitations}
Our threshold selection approach in Figure~\ref{fig:cyber-scatter} relies on estimations that require more rigorous examination. The absence of clear clustering suggests these thresholds may represent gradual transitions rather than abrupt boundaries. We also found that high standalone perplexity does not consistently indicate nonsensical text (see Appendix~\ref{appendix:categories}), challenging its reliability as a degeneration detector. For future work, we encourage exploring alternative evaluation methods, such as model-as-a-judge approaches~\cite{zheng2023judgingllmasajudgemtbenchchatbot}, to more accurately identify text degeneration. 

A methodological limitation worth addressing is the potential bias introduced by our prompt generation technique. Since some prompts originate from the Pile dataset, this artificially inflates certain sequence counts. Further studies incorporating manually crafted prompts would help quantify and mitigate this bias. 

Additionally, trying different model sizes, and including a wider set of prompts, from non-scientific domains without specific keywords would allow to state the limitations more clearly.

Finally, we note that our model uses the Pythia tokenizer, whereas Infinigram relies on the LLaMA-2 tokenizer. As a result, certain spans—especially verbatim sequences—may fail to align across models despite being present in the training data. We recommend performing indexing with the same tokenizer used at inference time to avoid such mismatches.

Our pipeline may serve as an additional tool for Training Data Attribution (TDA) investigations. We anticipate future research exploring the relationships between low-perplexity windows and sequences, as briefly discussed in Appendix~\ref{appendix:match}. Additionally, comparative analyses between our method and other state-of-the-art TDA approaches would be valuable for establishing best practices in this emerging field, alongside with efficiency measurements.

\section{Ethics statements}

Training data extraction is a threat to user privacy, as this can be used to find Personally Identifiable Information (PII) such as leaked passwords, address or contact information~\cite{brown2022doesmeanlanguagemodel}. We try to mitigate this in the following way. First, we work on a publicly available model, and use examples from Wikipedia, also publicly available. However, we acknowledge that the Pile dataset, which was used to train the Pythia models, contains copyrighted material~\cite{monology2021pileuncopyrighted}. Given these concerns, we advocate for future research to prioritize copyright-compliant datasets that respect creators' intellectual property rights while advancing our understanding of model behavior. On the other hand, our work contribute to training data transparency, and can help to detect copyright infringement. We also recall that our method requires to possess an indexing of the training data, which is not the case for the state-of-the-art models. We believe that the impact of this paper does not present direct major risks and encourage further work in this direction.

For transparency, we give an estimation of the CO$_2$ emitted by the computation. We used approximately 120 hours of GPU with an average consumption of \SI{250}{\watt}, and considering the CO$_2$ emissions per kilowatt-hour in the region we are located in to be \SI{38.30}{\gram\text{CO}_{2}\text{eq}\per\kilo\watt\hour}~\cite{lowcarbonpower2024}, this totals to $120 \times 0.25 \times 38.30 = \SI{1.1}{\kilo\gram\text{CO}_{2}\text{eq}}$.
%Switzerland \andrei{region where we are located; during anonymous review}

Finally, additional generative AI tools were used solely to assist with reformulating parts of the text and code for improved clarity and readability.

%\andrei{use it as more efficient source data attribution}

%\andrei{explore connections between low-perplexity sequences}

%\andrei{better understand memorization}

%\andrei{Explore efficiency of the training data usage}

\section*{Acknowledgments}
The authors are thankful to Alexander Sternfeld and Prof. Antoine Bosselut for their valuable input on the paper, and to the anonymous reviewers of ACL 2025 for their constructive comments. We additionally thank Prof. Bosselut for hosting Arthur Wuhrmann (AW) in his lab during the course of this work.
Andrei Kucharavy (ADK) and Anastasiia Kucherenko (AAK) are supported by the CYD Campus, armasuisse W+T, ARAMIS AR-CYD-C-025 grant.

\subsection*{Contributions} 
\begin{itemize}
\item Conceptualization: AAK, ADK; 
\item Methodology, Software, Data Curation, Visualization, and Writing - Original Draft: AW, ADK;
\item Investigation, Writing - Review \& Editing: AW, AAK, ADK; 
\item Supervision, Project Administration and Funding Acquisition: ADK.
\end{itemize}
%\andrei{This also determines the order of authors: AW, AAK, and ADK}   

%Each of the authors on this paper significantly contributed to the final results.
%\andrei{I would suggest using CRediT (https://credit.niso.org/) and remove it for now (review) and leaving it as part of the acknowledgements.}
%\begin{itemize}
%\item Anastasiia contributed to project scoping, designed experiments and contributed significantly to prompt extraction and writing.
%\item Andrei provided guidance and mentorship throughout the project, contributed to the writing and helped clearing out project scope. 
%\item Arthur built and ran the evaluation and text generation pipelines, contributed significantly to writing and analysis.
%\end{itemize}

\clearpage
\appendix
\section{Visualization of degeneration}
\label{appendix:degeneration}
\begin{figure}[!h]  % 't' places it at the top of the page
    \centering
    \includegraphics[width=0.9\columnwidth]{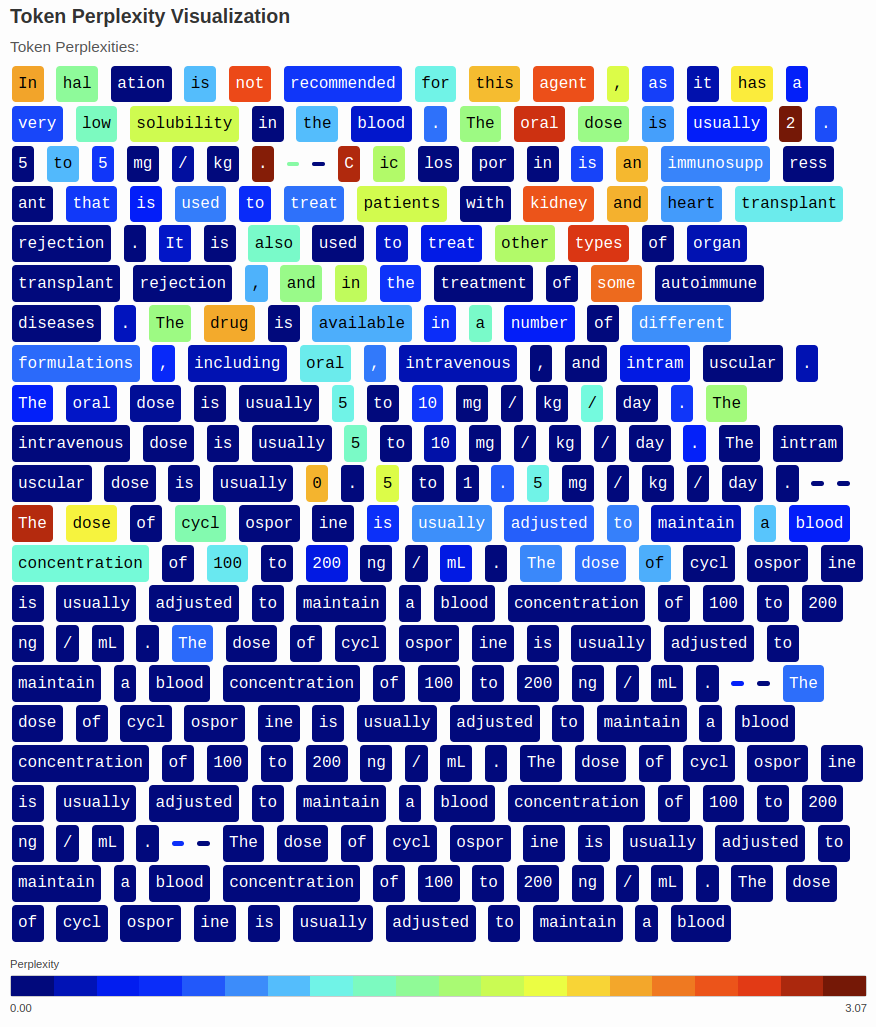}
    \caption{Example of the perplexities of an output that degenerates.}
    \label{fig:degeneration}
\end{figure}

While we did not include degeneration region in Fig.~\ref{fig:cyber-scatter}, we still encountered it during our experiments. Here, by degeneration, we refer to undesirable patterns in generated text, such as nonsensical or incoherent outputs, excessive repetition, and looping behaviors—where the model repeatedly generates the same tokens or phrases in a cyclic manner. Fig.~\ref{fig:degeneration} shows an example of it. This exclusion stemmed from two observations: the repetitive patterns extended beyond our window size parameters, and the degenerated text displayed surprisingly low standalone perplexity values. These findings highlight a limitation in using perplexity-based metrics alone for degeneration detection and suggest the need for complementary approaches.
\section{Examples of texts per category.}
Tab.~\ref{tab:example-table} presents examples of low-perplexity windows belonging to different categories. We also added example of high perplexities.
\label{appendix:categories}

\begin{table}[!h]
\centering
\small
\begin{tabularx}{\columnwidth}{p{2.5cm}X}
\toprule
\textbf{category} & \textbf{text} \\
\midrule
Frequently encountered text & -- synthetic cannabinoid. \\
 & -- a function that takes as input an  \\
 & -- Standards and Technology (NIST) \\
\addlinespace[6pt]
Memorization & -- alcohol, sugar, water, and \\
 & -- to the evaluation of a cryptographic \\
 & -- of information that is used to encrypt \\
\addlinespace[6pt]
Segmental replication & --  has been defined as "the study  \\ 
& -- used for PET and SPECT imaging \\ 
& -- understanding of the genetic basis of common  \\
\addlinespace[6pt]
Synthetic coherence & -- and genetics. fireball starts to form. The \\
 & -- the exchanged keys are computationally indistinguishable from \\
 & -- . Developmental genetics is also the \\
\addlinespace[6pt]
 High standalone perplexity ($\log_2(\hat{P}) > 12$) & -- random cipher digit stream (key \\
 & -- ion CN $-$ is also being removed \\
 & -- a sudden influx of fast neutrons \\
\bottomrule
\end{tabularx}
\caption{Examples of text fragments and their categories.}
\label{tab:example-table}
\end{table}
\section{Example of repetition.}
\label{appendix:repetition}
We show here an example of the model repeating the prompt. The repetition is highlighted in red. The end has been cut for better readability.

\begin{tcolorbox}[
  colback=white!10,
  colframe=titlebar,
  title=Example of Prompt Repetition,
  fonttitle=\bfseries\color{white},
  boxrule=1pt,
  arc=2pt
]

\textbf{Prompt:}
\begin{tcolorbox}[
  colback=promptbox,
  colframe=blue!20,
  boxrule=0.5pt,
  arc=2pt
]
There are a number of standards related to cryptography. Standard algorithms and protocols provide a focus for study; standards for popular applications attract a large amount of cryptanalysis.
\end{tcolorbox}

\vspace{0.3cm}

\textbf{Output:}
\begin{tcolorbox}[
  colback=outputbox,
  colframe=red!20,
  boxrule=0.5pt,
  arc=2pt
]
\textcolor{black}{Standards provide a focus for cryptanalysis.} 
\textcolor{repetition}{Standards for popular applications attract a large amount of cryptanalysis.} 
{\color{repetition}There are a number of standards related to cryptography.} 
{\color{repetition}Standard algorithms and protocols provide a focus for study; standards for popular applications attract a large amount of cryptanalysis. Standards provide a focus for cryptanalysis. Standards for popular applications attract [...]}
\end{tcolorbox}

\end{tcolorbox}

\section{Surrounding of sequences match}
\label{appendix:match}
When a sequence has a low number of matches (in the memorization part, see Sec.~\ref{sec:categories}) in the training data, one can look at the original document containing the sequence. %(this is a feature proposed by Infinigram~\cite{infinigram}). 
Below is an example. The part in bold has one exact match to the Pile, and the extract is shown below. 
\begin{tcolorbox}[
  colback=white!10,
  colframe=titlebar,
  title=Segment match ,
  fonttitle=\bfseries\color{white},
  boxrule=1pt,
  arc=2pt
]

\textbf{Prompt:}
\begin{tcolorbox}[
  colback=promptbox,
  colframe=blue!20,
  boxrule=0.5pt,
  arc=2pt
]
A message authentication code (MAC) is a short piece of information used to authenticate a message.\end{tcolorbox}

\vspace{0.3cm}

\textbf{Output:}
\begin{tcolorbox}[
  colback=outputbox,
  colframe=red!20,
  boxrule=0.5pt,
  arc=2pt
]
\textcolor{black}{The MAC is a function of a secret key and the message.}
\textcolor{black}{
The message is the information being protected. The key \textbf{is a secret known only to the sender and} receiver. The MAC is a function of the key and the message.
The MAC is a short piece of information used to authenticate a message. [...]}
\end{tcolorbox}

\vspace{0.5cm}
\textbf{Document matching:}
\begin{tcolorbox}[
  colback=docbox,
  colframe=green!20,
  boxrule=0.5pt,
  arc=2pt
]
\textcolor{black}{[...] HMAC is a well-known algorithm for generating a message authentication code (MAC) that can be used to verify the
integrity and authenticity of a message.\newline
This class requires Qt 4.3.0 or greater.\newline
To verify a message, the sender creates a MAC using a key, which \textbf{is a secret known only to the sender and} recipient,
and the content of the message. This MAC is then sent along with the message. The recipient then creates another MAC
using the shared key and the content of the message. If the two codes match, the message is verified. [...]}
\end{tcolorbox}
\end{tcolorbox}
\newpage 
The document comes from GitHub. Interestingly, while the low-perplexity window in itself does not refer to MAC, the matching document is talking about MAC. Although further investigation is required to assess this, it might indicate that the context between low-perplexity sequences that match to the training data is related to the original document.


\newpage

\begin{thebibliography}{31}
\providecommand{\natexlab}[1]{#1}

\bibitem[{Akyurek et~al.(2022)Akyurek, Bolukbasi, Liu, Xiong, Tenney, Andreas,
  and Guu}]{akyurek-etal-2022-towards}
Ekin Akyurek, Tolga Bolukbasi, Frederick Liu, Binbin Xiong, Ian Tenney, Jacob
  Andreas, and Kelvin Guu. 2022.
\newblock \href {https://doi.org/10.18653/v1/2022.findings-emnlp.180} {Towards
  tracing knowledge in language models back to the training data}.
\newblock In \emph{Findings of the Association for Computational Linguistics:
  EMNLP 2022}, pages 2429--2446, Abu Dhabi, United Arab Emirates. Association
  for Computational Linguistics.

\bibitem[{Al-Kaswan et~al.(2024)Al-Kaswan, Izadi, and van
  Deursen}]{Al_Kaswan_2024}
Ali Al-Kaswan, Maliheh Izadi, and Arie van Deursen. 2024.
\newblock \href {https://doi.org/10.1145/3597503.3639133} {Traces of
  memorisation in large language models for code}.
\newblock In \emph{Proceedings of the IEEE/ACM 46th International Conference on
  Software Engineering}, ICSE ’24, page 1–12. ACM.

\bibitem[{Bender et~al.(2021)Bender, Gebru, McMillan-Major, and
  Shmitchell}]{10.1145/3442188.3445922}
Emily~M. Bender, Timnit Gebru, Angelina McMillan-Major, and Shmargaret
  Shmitchell. 2021.
\newblock \href {https://doi.org/10.1145/3442188.3445922} {On the dangers of
  stochastic parrots: Can language models be too big?}
\newblock In \emph{Proceedings of the 2021 ACM Conference on Fairness,
  Accountability, and Transparency}, FAccT '21, page 610–623, New York, NY,
  USA. Association for Computing Machinery.

\bibitem[{Biderman et~al.(2023)Biderman, Schoelkopf, Anthony, Bradley, O'Brien,
  Hallahan, Khan, Purohit, Prashanth, Raff, Skowron, Sutawika, and van~der
  Wal}]{biderman2023pythiasuiteanalyzinglarge}
Stella Biderman, Hailey Schoelkopf, Quentin Anthony, Herbie Bradley, Kyle
  O'Brien, Eric Hallahan, Mohammad~Aflah Khan, Shivanshu Purohit, USVSN~Sai
  Prashanth, Edward Raff, Aviya Skowron, Lintang Sutawika, and Oskar van~der
  Wal. 2023.
\newblock \href {https://arxiv.org/abs/2304.01373} {Pythia: A suite for
  analyzing large language models across training and scaling}.
\newblock \emph{Preprint}, arXiv:2304.01373.

\bibitem[{Brown et~al.(2022)Brown, Lee, Mireshghallah, Shokri, and
  Tramèr}]{brown2022doesmeanlanguagemodel}
Hannah Brown, Katherine Lee, Fatemehsadat Mireshghallah, Reza Shokri, and
  Florian Tramèr. 2022.
\newblock \href {https://arxiv.org/abs/2202.05520} {What does it mean for a
  language model to preserve privacy?}
\newblock \emph{Preprint}, arXiv:2202.05520.

\bibitem[{Carlini et~al.(2023{\natexlab{a}})Carlini, Hayes, Nasr, Jagielski,
  Sehwag, Tramèr, Balle, Ippolito, and
  Wallace}]{carlini2023extractingtrainingdatadiffusion}
Nicholas Carlini, Jamie Hayes, Milad Nasr, Matthew Jagielski, Vikash Sehwag,
  Florian Tramèr, Borja Balle, Daphne Ippolito, and Eric Wallace.
  2023{\natexlab{a}}.
\newblock \href {https://arxiv.org/abs/2301.13188} {Extracting training data
  from diffusion models}.
\newblock \emph{Preprint}, arXiv:2301.13188.

\bibitem[{Carlini et~al.(2023{\natexlab{b}})Carlini, Ippolito, Jagielski, Lee,
  Tramer, and Zhang}]{carlini2023quantifyingmemorizationneurallanguage}
Nicholas Carlini, Daphne Ippolito, Matthew Jagielski, Katherine Lee, Florian
  Tramer, and Chiyuan Zhang. 2023{\natexlab{b}}.
\newblock \href {https://arxiv.org/abs/2202.07646} {Quantifying memorization
  across neural language models}.
\newblock \emph{Preprint}, arXiv:2202.07646.

\bibitem[{Carlini et~al.(2021)Carlini, Tramer, Wallace, Jagielski,
  Herbert-Voss, Lee, Roberts, Brown, Song, Erlingsson, Oprea, and
  Raffel}]{carlini2021extractingtrainingdatalarge}
Nicholas Carlini, Florian Tramer, Eric Wallace, Matthew Jagielski, Ariel
  Herbert-Voss, Katherine Lee, Adam Roberts, Tom Brown, Dawn Song, Ulfar
  Erlingsson, Alina Oprea, and Colin Raffel. 2021.
\newblock \href {https://arxiv.org/abs/2012.07805} {Extracting training data
  from large language models}.
\newblock \emph{Preprint}, arXiv:2012.07805.

\bibitem[{Chang et~al.(2024)Chang, Rajagopal, Bolukbasi, Dixon, and
  Tenney}]{chang2024scalableinfluencefacttracing}
Tyler~A. Chang, Dheeraj Rajagopal, Tolga Bolukbasi, Lucas Dixon, and Ian
  Tenney. 2024.
\newblock \href {https://arxiv.org/abs/2410.17413} {Scalable influence and fact
  tracing for large language model pretraining}.
\newblock \emph{Preprint}, arXiv:2410.17413.

\bibitem[{Cheng et~al.(2025)Cheng, Bae, Bullock, and
  Kristofferson}]{cheng2025trainingdataattributiontda}
Deric Cheng, Juhan Bae, Justin Bullock, and David Kristofferson. 2025.
\newblock \href {https://arxiv.org/abs/2501.12642} {Training data attribution
  (tda): Examining its adoption \& use cases}.
\newblock \emph{Preprint}, arXiv:2501.12642.

\bibitem[{Deguchi et~al.(2025)Deguchi, Kamoda, Matsushita, Taguchi, Suenaga,
  Waga, and Yokoi}]{deguchi2025softmatchasoftfastpattern}
Hiroyuki Deguchi, Go~Kamoda, Yusuke Matsushita, Chihiro Taguchi, Kohei Suenaga,
  Masaki Waga, and Sho Yokoi. 2025.
\newblock \href {https://arxiv.org/abs/2503.03703} {Softmatcha: A soft and fast
  pattern matcher for billion-scale corpus searches}.
\newblock \emph{Preprint}, arXiv:2503.03703.

\bibitem[{Feldman and Zhang(2020)}]{10.5555/3495724.3495966}
Vitaly Feldman and Chiyuan Zhang. 2020.
\newblock What neural networks memorize and why: discovering the long tail via
  influence estimation.
\newblock In \emph{Proceedings of the 34th International Conference on Neural
  Information Processing Systems}, NIPS '20, Red Hook, NY, USA. Curran
  Associates Inc.

\bibitem[{Gao et~al.(2019)Gao, He, Tan, Qin, Wang, and
  Liu}]{gao2019representationdegenerationproblemtraining}
Jun Gao, Di~He, Xu~Tan, Tao Qin, Liwei Wang, and Tie-Yan Liu. 2019.
\newblock \href {https://arxiv.org/abs/1907.12009} {Representation degeneration
  problem in training natural language generation models}.
\newblock \emph{Preprint}, arXiv:1907.12009.

\bibitem[{Gao et~al.(2020)Gao, Biderman, Black, Golding, Hoppe, Foster, Phang,
  He, Thite, Nabeshima, Presser, and Leahy}]{gao2020pile800gbdatasetdiverse}
Leo Gao, Stella Biderman, Sid Black, Laurence Golding, Travis Hoppe, Charles
  Foster, Jason Phang, Horace He, Anish Thite, Noa Nabeshima, Shawn Presser,
  and Connor Leahy. 2020.
\newblock \href {https://arxiv.org/abs/2101.00027} {The pile: An 800gb dataset
  of diverse text for language modeling}.
\newblock \emph{Preprint}, arXiv:2101.00027.

\bibitem[{Gonen et~al.(2024)Gonen, Iyer, Blevins, Smith, and
  Zettlemoyer}]{gonen2024demystifyingpromptslanguagemodels}
Hila Gonen, Srini Iyer, Terra Blevins, Noah~A. Smith, and Luke Zettlemoyer.
  2024.
\newblock \href {https://arxiv.org/abs/2212.04037} {Demystifying prompts in
  language models via perplexity estimation}.
\newblock \emph{Preprint}, arXiv:2212.04037.

\bibitem[{Gormley and Tong(2015)}]{gormley2015elasticsearch}
Clinton Gormley and Zachary Tong. 2015.
\newblock \href
  {https://www.oreilly.com/library/view/elasticsearch-the-definitive/9781449358532/}
  {\emph{Elasticsearch: The Definitive Guide}}.
\newblock O'Reilly Media.

\bibitem[{Guu et~al.(2023)Guu, Webson, Pavlick, Dixon, Tenney, and
  Bolukbasi}]{guu2023simfluencemodelinginfluenceindividual}
Kelvin Guu, Albert Webson, Ellie Pavlick, Lucas Dixon, Ian Tenney, and Tolga
  Bolukbasi. 2023.
\newblock \href {https://arxiv.org/abs/2303.08114} {Simfluence: Modeling the
  influence of individual training examples by simulating training runs}.
\newblock \emph{Preprint}, arXiv:2303.08114.

\bibitem[{Khandelwal et~al.(2020)Khandelwal, Levy, Jurafsky, Zettlemoyer, and
  Lewis}]{khandelwal2020generalizationmemorizationnearestneighbor}
Urvashi Khandelwal, Omer Levy, Dan Jurafsky, Luke Zettlemoyer, and Mike Lewis.
  2020.
\newblock \href {https://arxiv.org/abs/1911.00172} {Generalization through
  memorization: Nearest neighbor language models}.
\newblock \emph{Preprint}, arXiv:1911.00172.

\bibitem[{Kwon et~al.(2023)Kwon, Wu, Wu, and
  Zou}]{DBLP:journals/corr/abs-2310-00902}
Yongchan Kwon, Eric Wu, Kevin Wu, and James Zou. 2023.
\newblock \href {https://doi.org/10.48550/ARXIV.2310.00902} {Datainf:
  Efficiently estimating data influence in lora-tuned llms and diffusion
  models}.
\newblock \emph{CoRR}, abs/2310.00902.

\bibitem[{Liang et~al.(2024)Liang, Zhang, Wu, Lepp, Ji, Zhao, Cao, Liu, He,
  Huang, Yang, Potts, Manning, and Zou}]{liang2024mappingincreasingusellms}
Weixin Liang, Yaohui Zhang, Zhengxuan Wu, Haley Lepp, Wenlong Ji, Xuandong
  Zhao, Hancheng Cao, Sheng Liu, Siyu He, Zhi Huang, Diyi Yang, Christopher
  Potts, Christopher~D Manning, and James~Y. Zou. 2024.
\newblock \href {https://arxiv.org/abs/2404.01268} {Mapping the increasing use
  of llms in scientific papers}.
\newblock \emph{Preprint}, arXiv:2404.01268.

\bibitem[{Liu et~al.(2025{\natexlab{a}})Liu, Blanton, Elazar, Min, Chen,
  Chheda-Kothary, Tran, Bischoff, Marsh, Schmitz, Trier, Sarnat, James,
  Borchardt, Kuehl, Cheng, Farley, Sreeram, Anderson, Albright, Schoenick,
  Soldaini, Groeneveld, Pang, Koh, Smith, Lebrecht, Choi, Hajishirzi, Farhadi,
  and Dodge}]{olmotrace}
Jiacheng Liu, Taylor Blanton, Yanai Elazar, Sewon Min, YenSung Chen, Arnavi
  Chheda-Kothary, Huy Tran, Byron Bischoff, Eric Marsh, Michael Schmitz,
  Cassidy Trier, Aaron Sarnat, Jenna James, Jon Borchardt, Bailey Kuehl, Evie
  Cheng, Karen Farley, Sruthi Sreeram, Taira Anderson, and 12 others.
  2025{\natexlab{a}}.
\newblock \href {https://arxiv.org/abs/2504.07096} {Olmotrace: Tracing language
  model outputs back to trillions of training tokens}.
\newblock \emph{Preprint}, arXiv:2504.07096.

\bibitem[{Liu et~al.(2025{\natexlab{b}})Liu, Min, Zettlemoyer, Choi, and
  Hajishirzi}]{infinigram}
Jiacheng Liu, Sewon Min, Luke Zettlemoyer, Yejin Choi, and Hannaneh Hajishirzi.
  2025{\natexlab{b}}.
\newblock \href {https://arxiv.org/abs/2401.17377} {Infini-gram: Scaling
  unbounded n-gram language models to a trillion tokens}.
\newblock \emph{Preprint}, arXiv:2401.17377.

\bibitem[{McCoy et~al.(2023)McCoy, Smolensky, Linzen, Gao, and
  Celikyilmaz}]{mccoy-etal-2023-much}
R.~Thomas McCoy, Paul Smolensky, Tal Linzen, Jianfeng Gao, and Asli
  Celikyilmaz. 2023.
\newblock \href {https://doi.org/10.1162/tacl_a_00567} {How much do language
  models copy from their training data? evaluating linguistic novelty in text
  generation using {RAVEN}}.
\newblock \emph{Transactions of the Association for Computational Linguistics},
  11:652--670.

\bibitem[{Merrill et~al.(2024)Merrill, Smith, and
  Elazar}]{merrill2024evaluatingngramnoveltylanguage}
William Merrill, Noah~A. Smith, and Yanai Elazar. 2024.
\newblock \href {https://arxiv.org/abs/2406.13069} {Evaluating $n$-gram novelty
  of language models using rusty-dawg}.
\newblock \emph{Preprint}, arXiv:2406.13069.

\bibitem[{Monology(2021)}]{monology2021pileuncopyrighted}
Monology. 2021.
\newblock Pile uncopyrighted.
\newblock \url{https://huggingface.co/datasets/monology/pile-uncopyrighted}.
\newblock Accessed: May 17, 2025.

\bibitem[{Pan et~al.(2025)Pan, Shi, Zhao, and Ma}]{detecting_Yijun}
Yijun Pan, Taiwei Shi, Jieyu Zhao, and Jiaqi Ma. 2025.
\newblock \href {https://doi.org/10.48550/arXiv.2502.11411} {Detecting and
  filtering unsafe training data via data attribution}.

\bibitem[{Power(2024)}]{lowcarbonpower2024}
Low-Carbon Power. 2024.
\newblock \href {https://lowcarbonpower.org/region/Switzerland} {Carbon
  intensity of electricity in switzerland}.
\newblock Accessed: May 17, 2025.

\bibitem[{Prashanth et~al.(2025)Prashanth, Deng, O'Brien, V, Khan, Borkar,
  Choquette-Choo, Fuehne, Biderman, Ke, Lee, and
  Saphra}]{prashanth2025recitereconstructrecollectmemorization}
USVSN~Sai Prashanth, Alvin Deng, Kyle O'Brien, Jyothir~S V, Mohammad~Aflah
  Khan, Jaydeep Borkar, Christopher~A. Choquette-Choo, Jacob~Ray Fuehne, Stella
  Biderman, Tracy Ke, Katherine Lee, and Naomi Saphra. 2025.
\newblock \href {https://arxiv.org/abs/2406.17746} {Recite, reconstruct,
  recollect: Memorization in lms as a multifaceted phenomenon}.
\newblock \emph{Preprint}, arXiv:2406.17746.

\bibitem[{Review(2024)}]{MIT2024LLMs}
MIT~Technology Review. 2024.
\newblock \href
  {https://www.technologyreview.com/2024/03/04/1089403/large-language-models-amazing-but-nobody-knows-why/}
  {Large language models can do jaw-dropping things. but nobody knows exactly
  why.}
\newblock Accessed: 2025-05-18.

\bibitem[{Wu et~al.(2024)Wu, Pang, Shen, and
  Cheng}]{wu-etal-2024-enhancing-training}
Kangxi Wu, Liang Pang, Huawei Shen, and Xueqi Cheng. 2024.
\newblock \href {https://doi.org/10.18653/v1/2024.emnlp-main.782} {Enhancing
  training data attribution for large language models with fitting error
  consideration}.
\newblock In \emph{Proceedings of the 2024 Conference on Empirical Methods in
  Natural Language Processing}, pages 14131--14143, Miami, Florida, USA.
  Association for Computational Linguistics.

\bibitem[{Zheng et~al.(2023)Zheng, Chiang, Sheng, Zhuang, Wu, Zhuang, Lin, Li,
  Li, Xing, Zhang, Gonzalez, and
  Stoica}]{zheng2023judgingllmasajudgemtbenchchatbot}
Lianmin Zheng, Wei-Lin Chiang, Ying Sheng, Siyuan Zhuang, Zhanghao Wu, Yonghao
  Zhuang, Zi~Lin, Zhuohan Li, Dacheng Li, Eric~P. Xing, Hao Zhang, Joseph~E.
  Gonzalez, and Ion Stoica. 2023.
\newblock \href {https://arxiv.org/abs/2306.05685} {Judging llm-as-a-judge with
  mt-bench and chatbot arena}.
\newblock \emph{Preprint}, arXiv:2306.05685.

\end{thebibliography}
\end{document}